# Investigate the Correlation of Breast Cancer Dataset using Different Clustering Technique

Somenath Chakraborty
School of Computing Sciences and Computer Engineering
The University of southern Mississippi
Hattiesburg, USA
Somenath.Chakraborty@usm.edu

Beddhu Murali
School of Computing Sciences and Computer Engineering
The University of southern Mississippi
Hattiesburg, USA
Bedhu.Murali@usm.edu

*Abstract*— The objectives of this paper are to explore ways to analyze breast cancer dataset in the context of unsupervised learning without prior training model. The paper investigates different ways of clustering techniques as well as preprocessing. This in-depth analysis builds the footprint which can further use for designing a most robust and accurate medical prognosis system. This paper also give emphasis on correlations of data points with different standard benchmark techniques.

## I. INTRODUCTION

Clustering is the task of dividing the population or data points into several groups such that data points in the same groups are more like other data points in the same group than those in other groups. It is the technique to classify objects or cases into relative groups called clusters. In simple words, the aim is to segregate groups with similar traits and assign them into clusters. Clustering is an unsupervised most common exploratory data analysis mechanism where there are no predefined classes exists, so we need to analyze the dataset to find different kinds of similarity among them. It is basically two types,

i)Hard Clustering: In hard clustering, each data point either belongs to a cluster completely or not.

ii)Soft Clustering: In soft clustering, instead of putting each data point into a separate cluster, a probability or likelihood of that data point to be in those clusters is assigned.

To make the existing model more accurate and robust we need class level and target definition of a model need to be very precise, which cannot be possible without accurate clustering. Our aim is to investigate the database correlation in a more comprehensive way so that it can segregate the dataset with utmost accuracy.

Cancer is a group of diseases involving abnormal cell growth with the potential to invade or spread to other parts of the body.

Breast cancer is one of the major kinds of cancer that develops from breast tissue. Signs of breast cancer may include a lump in the breast, a change in breast shape, dimpling of the skin, fluid coming from the nipple, a newly inverted nipple, or a red or scaly patch of skin.

According to the US's leading national public health institute Centers for Disease Control and Prevention (CDC), Breast cancer is the second most common cancer among women in the United States.

In 2016, the latest year for which incidence data are available, 245,299 new cases of Female Breast Cancer were reported, and 41,487 women died of Female Breast Cancer in the United States. For every 100,000 women, 124 new Female Breast Cancer cases were reported and 20 died of cancer.

Here in this paper the main objective is to analysis the Breast Cancer dataset using Clustering Technique.

## II. PREVIOUS WORK

In this section, the paper describes about existing research for Breast cancer medical prognosis model.

S. Muthukumaran et al.[1], Nepomuceno et al.[2], Doruk Bozdag et al.[3] use Biclustering approach. Sangeetha V [4] et al. uses genetic algorithm and Spiking Neural Networks to generate cluster in their breast cancer dataset. Their integration of genetic algorithm and Spiking Neural Network helps to develop the prognosis system. T. Padhi [5] et al., D. Verma et al. [7] uses weka Software tool to get the cluster for breast cancer dataset. L. M. Naeni et al. [6], Taosheng Xu et al.[8] uses gene expression and biomarker to investigate gene expression profiling to cluster the cancer dataset. A. L. Fijri et al. [9], P. H. S. Coutinho et al. [10] describe in their paper some kinds of Fuzzy analysis to cluster the dataset. D. Wu, L. Sheng et al. [11] and Chen D et al. [12] describe in their paper an ensemble algorithm for clustering cancer data. This algorithm is based on the k-medoids or partitioning around medoids (PAM) algorithm.

All this existing research mainly focus on predicting the class in known perspective rather than investigation of the fundamental unsupervised mechanism to segregate the datapoint in the dataset.

Investigation of the dataset in unsupervised way is very import in terms of designing more accurate model for data labeling which further can be used by classification and prediction problem.

### III. METHODOLOGY

#### A. Data Source

The dataset we used in our paper was obtained from the University of Wisconsin Hospitals, Madison from Dr. William H. Wolberg. The data was donated on July 15th, 1992 and has an integer attribute characteristic. It is also a multivariate data set. This dataset has been used by Wolberg in 1990 and Zhang in 1992. This dataset arrives periodically as Dr. Wolberg does his clinical reports. In total he recorded 699 patients with 11 attributes. The table shows the attributes he used:

Table 1: Details of Attributes of Breast Cancer Data

| # Attributes | Domain |
| --- | --- |
| 1. Sample code number | Id Number |
| 2. Clump Thickness | 1-10 |
| 3. Uniformity of Cell Size | 1-10 |
| 4. Uniformity of Cell Shape | 1-10 |
| 5. Marginal Adhesion | 1-10 |
| 6. Single Epithelial Cell Size | 1-10 |
| 7. Bare Nuclei | 1-10 |
| 8. Bland Chromatin | 1-10 |
| 9. Normal Nucleoli | 1-10 |
| 10. Mitoses | 1-10 |
| 11. Class | 2 for benign, 4 for malignant |

From 2 to 10 for each instance are rated between 1 to 10, respectively. The Class exhibits whether a patient is benign or malignant. This data had 16 missing attribute values. The dataset consisted of 458 benign which is 65.5% of the dataset and 241 malignant which is also 34.5% of the dataset.

#### B. Data Preprocessing

The data is downloaded in .csv format. In the csv extension, we delete all the 16 rows that had the missing values. After that the csv extension was converted to a Weka friendly file which is Attribute Relation Data Format (ARFF)extension. Then the data gone through a lot of preprocessing.

No. of Instances in the dataset before Preprocessing = 699

No. of Instances in the dataset after Preprocessing = 683

Then we use Weka software to further normalize the dataset using min-max normalization method so that all the feature values are in [0, 1]. Clustering is an unsupervised learning method, so only feature values are used for clustering. This means we do not normalize the category label which is the last column in the dataset.

First, we remove the ID number. Then we use Weka to further preprocess the dataset.

Then we investigate the dataset to understand correlation among the datapoints using Hopkins Statistic Index to understand whether the dataset has good clustering tendency or not.

Then we perform different sort of clustering technique both using Weka Software Tools and R programing language

### IV. RESULT

In our Breast Cancer data, we got Hopkins Statistic Index = 0.7997314 which indicates the dataset is highly clustered.

K-means Clustering Analysis Results:

Table 2: Cluster Instances after K-means

| Cluster Instances | | |
| --- | --- | --- |
| Benign | 402 | 59% |
| Malignant | 281 | 41% |

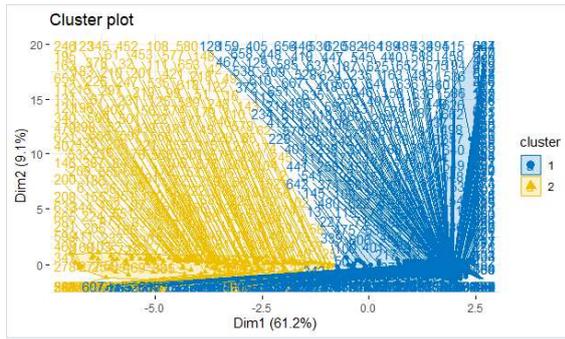

Fig.1 K-means cluster visualization

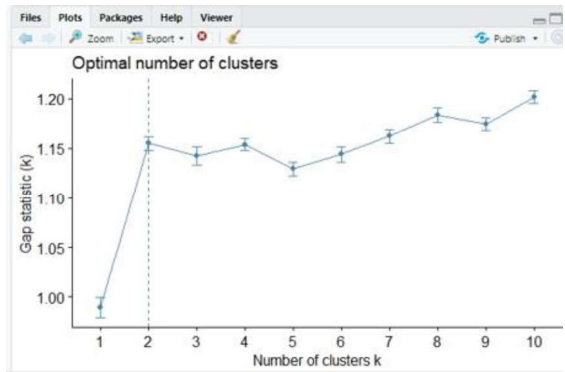

Fig.2 Box plot Which clearly shows the dataset has two optimal clusters.

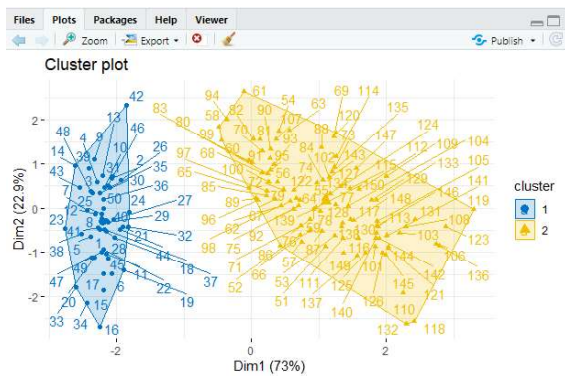

Fig.3 K-means clustering (Changing the visualization way, and some parameters)

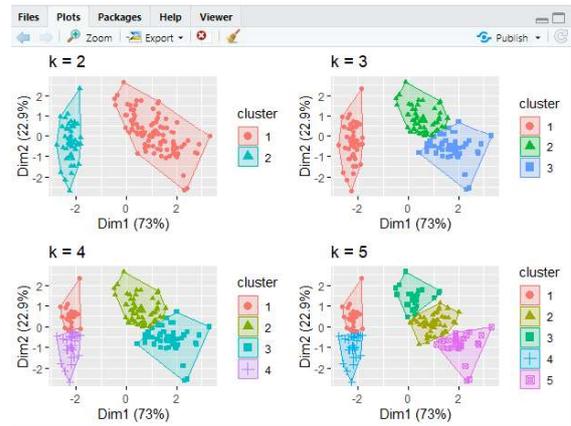

Fig.4 K-means for different values of k

K-medoids clustering or Partitioning Around Medoids (PAM) results:

An alternative to k-means clustering is the K-medoids clustering or PAM (Partitioning Around Medoids, Kaufman & Rousseeuw, 1990), which is less sensitive to outliers compared to k-means.

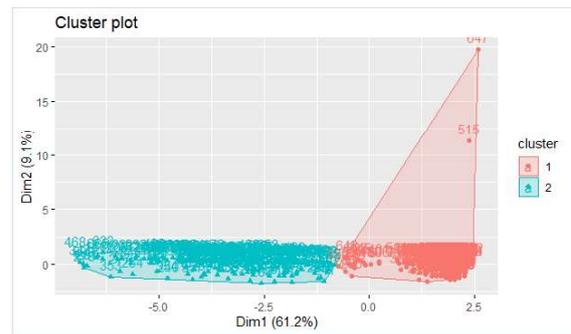

Fig.5 K-medoids clustering or PAM (Partitioning Around Medoids)

Silhouette Analysis Using R on Breast Cancer Dataset:

Silhouette analysis allows you to calculate how similar each observation is with the cluster it is assigned relative to other clusters. This metric (silhouette width) ranges from -1 to 1 for each observation in your data and can be interpreted as follows:

i)      Values close to 1 suggest that the observation is well matched to the assigned cluster.

ii)     Values close to 0 suggest that the observation is borderline matched between two clusters.

iii) Values close to -1 suggest that the observations may be assigned to the wrong cluster.

In our paper we got the metric (silhouette width) value = 0.57 which clearly shows that the Breast Cancer Dataset well matched to the assigned cluster.

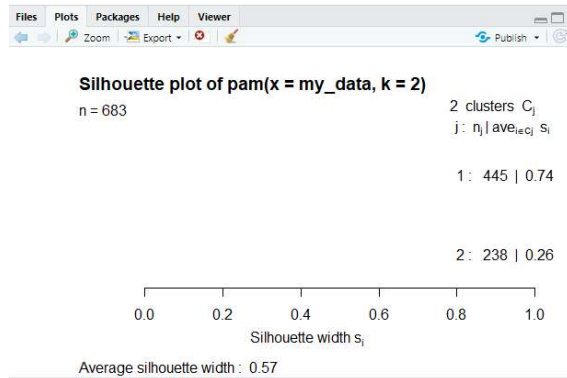

Fig.6 Silhouette Plot of K-medoids or PAM

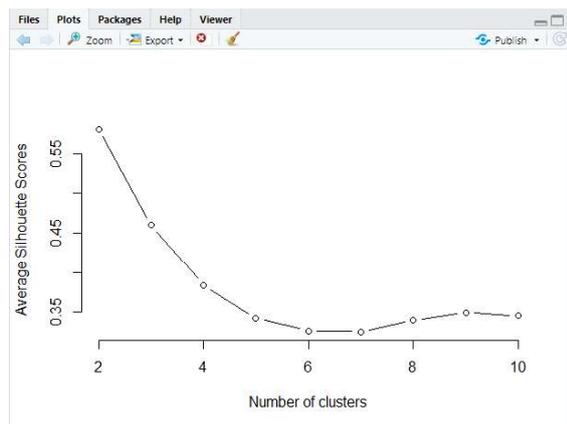

Fig.7 Average Silhouette Score plot of K-means clustering.

Average Silhouette Score plot of K-means clustering where K varies from 2 to 10 and the Highest Silhouette value is 0.582 which shows that the Breast Cancer Dataset well matched to the assigned cluster, when cluster size =2.

V. DISCUSSION AND FUTURE WORK

In this paper we use Weka machine learning software and R Programming language and R Visual studio to experiment with our breast cancer dataset. We use different clustering analysis to investigate proper correlation in our Breast cancer dataset. This is unsupervised learning paradigm where no pretrain model and label is needed.

In future we will develop a classification model using this unsupervised learning model as a featured trained system which can fed to the system to predict the values of model.

ACKNOWLEDGMENT

The authors acknowledge HPC at The University of Southern Mississippi supported by the National Science Foundation under the Major Research Instrumentation (MRI) program via Grant # ACI 1626217 and Prof. Chaoyang Zhang for his support and advice.